\documentclass[10pt,twocolumn,letterpaper]{article}

\usepackage{cvpr}              % To produce the CAMERA-READY version
\usepackage{graphicx}
\usepackage{amsmath}
\usepackage{amssymb}
\usepackage{booktabs}
\usepackage{multirow}

\usepackage[pagebackref,breaklinks,colorlinks]{hyperref}

\usepackage[capitalize]{cleveref}
\crefname{section}{Sec.}{Secs.}
\Crefname{section}{Section}{Sections}
\Crefname{table}{Table}{Tables}
\crefname{table}{Tab.}{Tabs.}
\newcommand{\comment}[1]{}

\def\cvprPaperID{*****} % *** Enter the CVPR Paper ID here
\def\confName{CVPR}
\def\confYear{2022}

\begin{document}

\title{Masked Image Modeling Advances 3D Medical Image Analysis}

% \comment{
\author{Zekai Chen\qquad Devansh Agarwal\qquad Kshitij Aggarwal\qquad Wiem Safta\\
Samit Hirawat\qquad Venkat Sethuraman\qquad Mariann Micsinai Balan\qquad Kevin Brown\\
Bristol Myers Squibb\\
{\tt\small \{zekai.chen, devansh.agarwal, kshitij.aggarwal, wiem.safta2, kevin.brown\}@bms.com}
% \and
}
% }
\maketitle

%%%%%%%%% ABSTRACT
\begin{abstract}
Recently, masked image modeling (MIM) has gained considerable attention due to its capacity to learn from vast amounts of unlabeled data and has been demonstrated to be effective on a wide variety of vision tasks involving natural images. Meanwhile, the potential of self-supervised learning in modeling 3D medical images is anticipated to be immense due to the high quantities of unlabeled images, and the expense and difficulty of quality labels. However, MIM's applicability to medical images remains uncertain. In this paper, we demonstrate that masked image modeling approaches can also advance 3D medical images analysis in addition to natural images. We study how masked image modeling strategies leverage performance from the viewpoints of 3D medical image segmentation as a representative downstream task: i) when compared to naive contrastive learning, masked image modeling approaches accelerate the convergence of supervised training even faster (1.40$\times$) and ultimately produce a higher dice score; ii) predicting raw voxel values with a high masking ratio and a relatively smaller patch size is non-trivial self-supervised pretext-task for medical images modeling; iii) a lightweight decoder or projection head design for reconstruction is powerful for masked image modeling on 3D medical images which speeds up training and reduce cost; iv) finally, we also investigate the effectiveness of MIM methods under different practical scenarios where different image resolutions and labeled data ratios are applied. 
% Self-supervised learning using masked image modeling (MIM) has gained considerable attention recently due to its capacity to learn from vast amounts of unlabeled data and has been demonstrated to be effective on a wide variety of vision tasks. At the same time, the potential of self-supervised learning in modeling 3D medical images is immense due to the high quantities of unlabeled images, and the expense and difficulty of quality labels. 

\end{abstract}

%%%%%%%%% BODY TEXT
\section{Introduction}
\label{sec:intro}

The demand for deep neural networks that conduct analysis tasks on 3D medical image data has expanded dramatically in recent years as a result of technological advances in deep learning and hardware compute capabilities. 3D medical volumetric images show a lot of potential in healthcare, where it can help increase the speed and accuracy of diagnosing patient conditions. For instance, properly and swiftly discovering and measuring tumor lesions from MRI/CT scans would be critical to disease prevention, early detection and treatment plan optimization, and would also spur the development of more successful clinical applications that would ultimately improve patients' lives~\cite{Bera2021PredictingCO}. However, the high expense of expert annotation frequently stymies attempts to leverage advances in clinical outcomes using deep learning approaches. Annotations of 3D medical images at scale by radiologists are limited, expensive, and time-consuming to produce. Another barrier in 3D medical imaging is data volume, which is driven by the increased 3D image dimensionality and resolution, resulting in significant processing complexity. As a consequence, training deep learning models on 3D medical images from random initialization necessitates burdensome compute and data requirements.
% As a consequence, in some cases, such as clinical study design, effectively integrating radiomics endpoint information with other bio-marker data for other downstream tasks such as tumor burden assessment, overall survival prediction, and so on can be extremely difficult.
\comment{
\begin{figure}
  \centering
  \includegraphics[width=\linewidth]{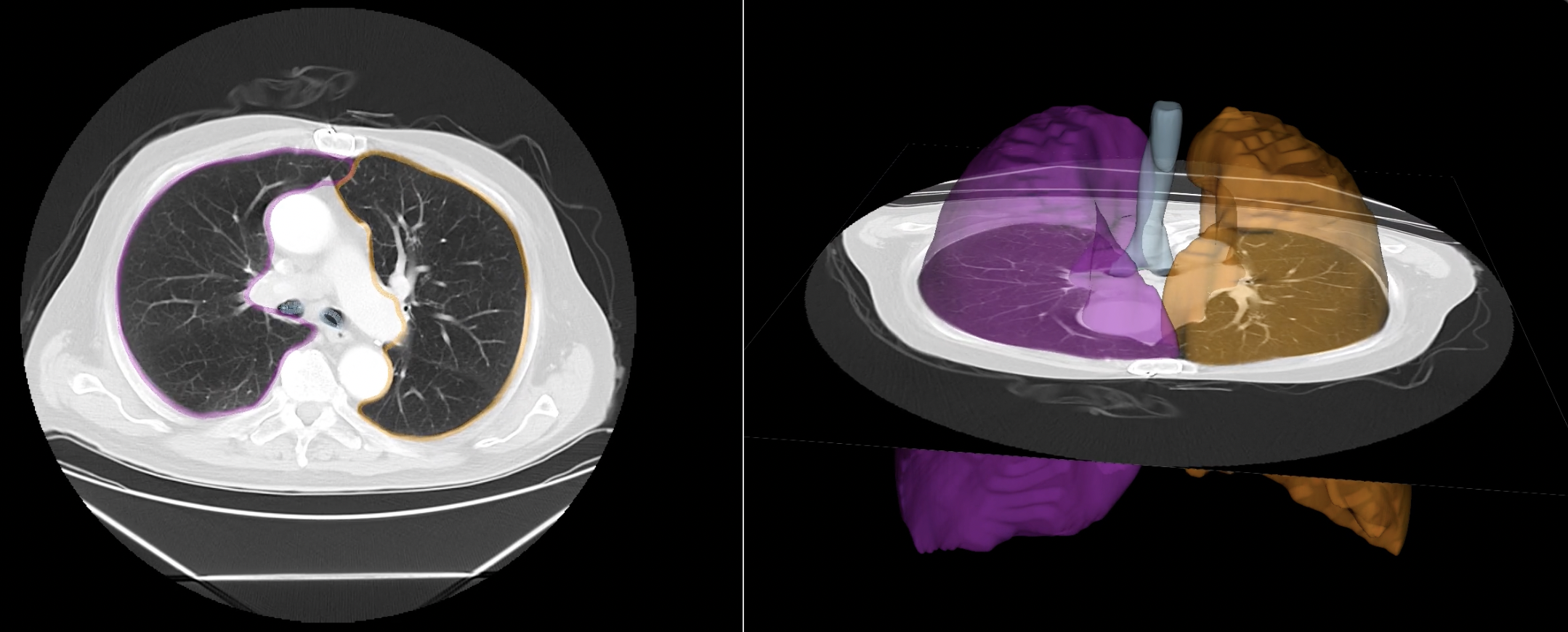}
  \caption{An example of one CT 3D scan of lung with lesions (right) and one slice at the middle depth (left). }
  \label{fig:intro}
\end{figure}
}

%-------------------------------------------------------------------------
% \comment{
\begin{figure*}
  \centering
  \begin{subfigure}[b]{0.2\linewidth}
    % \fbox{\rule{0pt}{2in} \rule{.9\linewidth}{0pt}}
    \includegraphics[width=\linewidth]{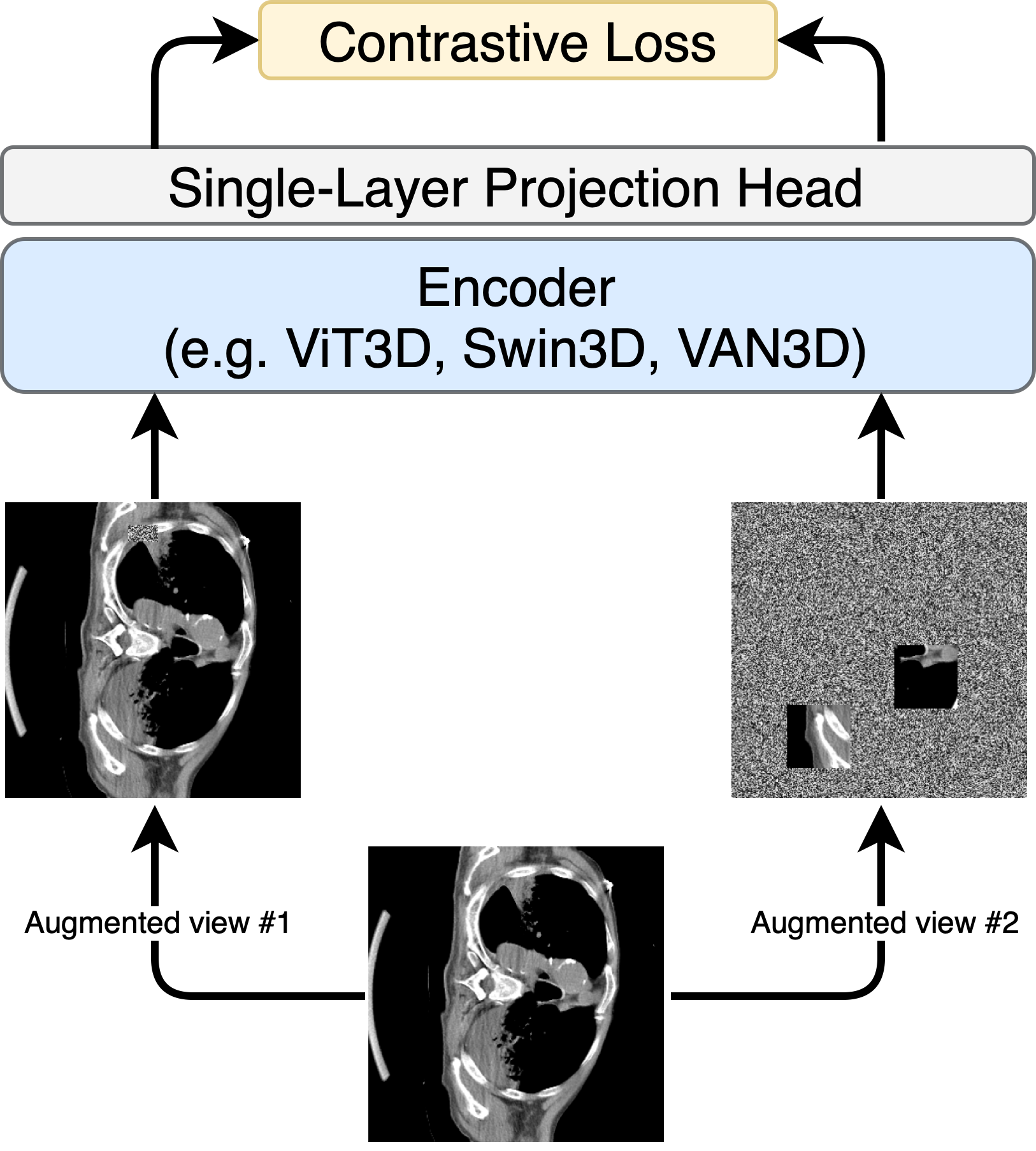}
    \caption{The SimCLR architecture.}
    \label{fig:simclr}
  \end{subfigure}
  \hfill
  \begin{subfigure}[b]{0.38\linewidth}
   % \fbox{\rule{0pt}{2in} \rule{.9\linewidth}{0pt}}
   \includegraphics[width=\linewidth]{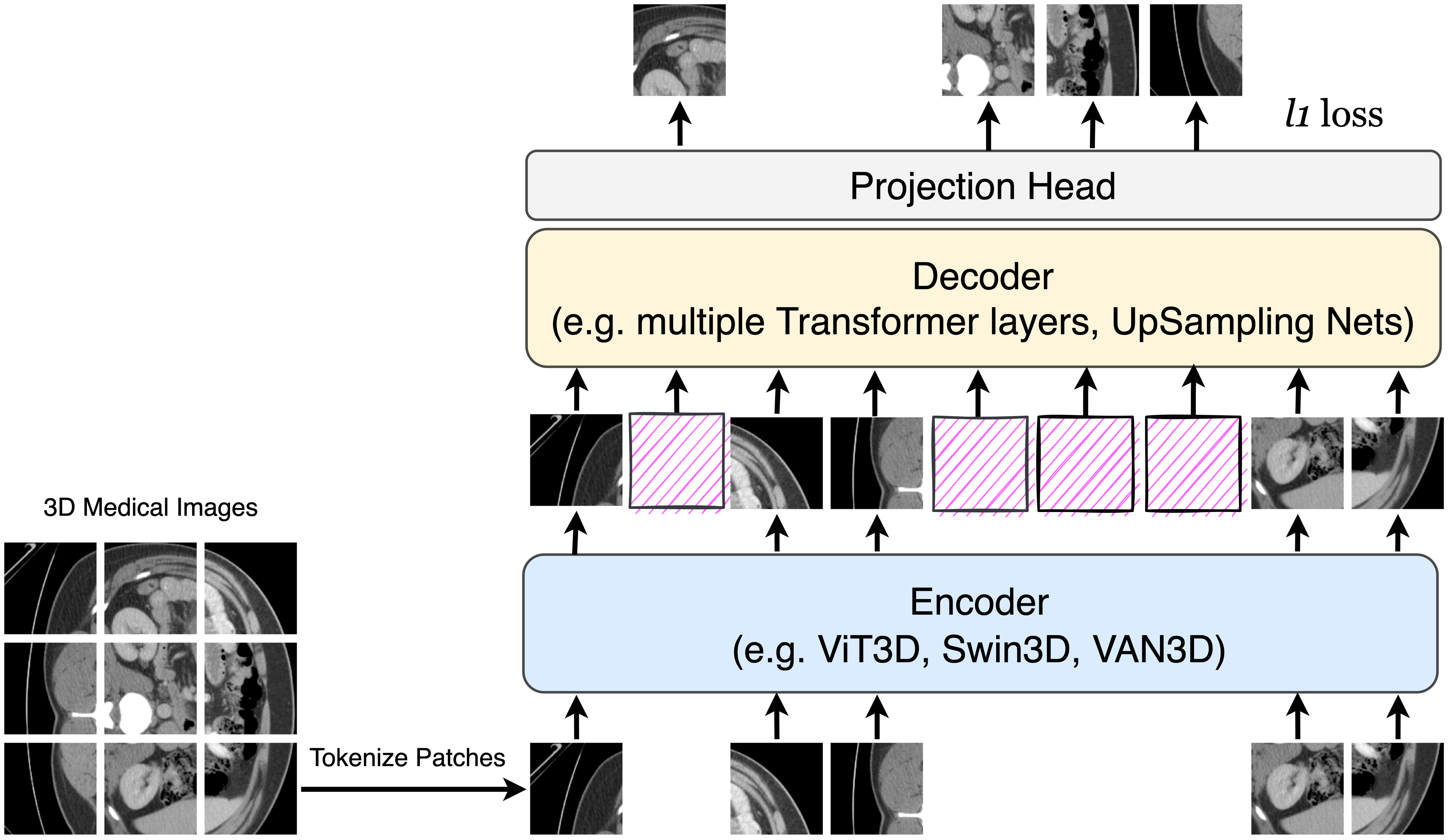}
    \caption{The MAE architecture.}
    \label{fig:mae}
  \end{subfigure}
  \hfill
  \begin{subfigure}[b]{0.38\linewidth}
    % \fbox{\rule{0pt}{2in} \rule{.9\linewidth}{0pt}}
    \includegraphics[width=\linewidth]{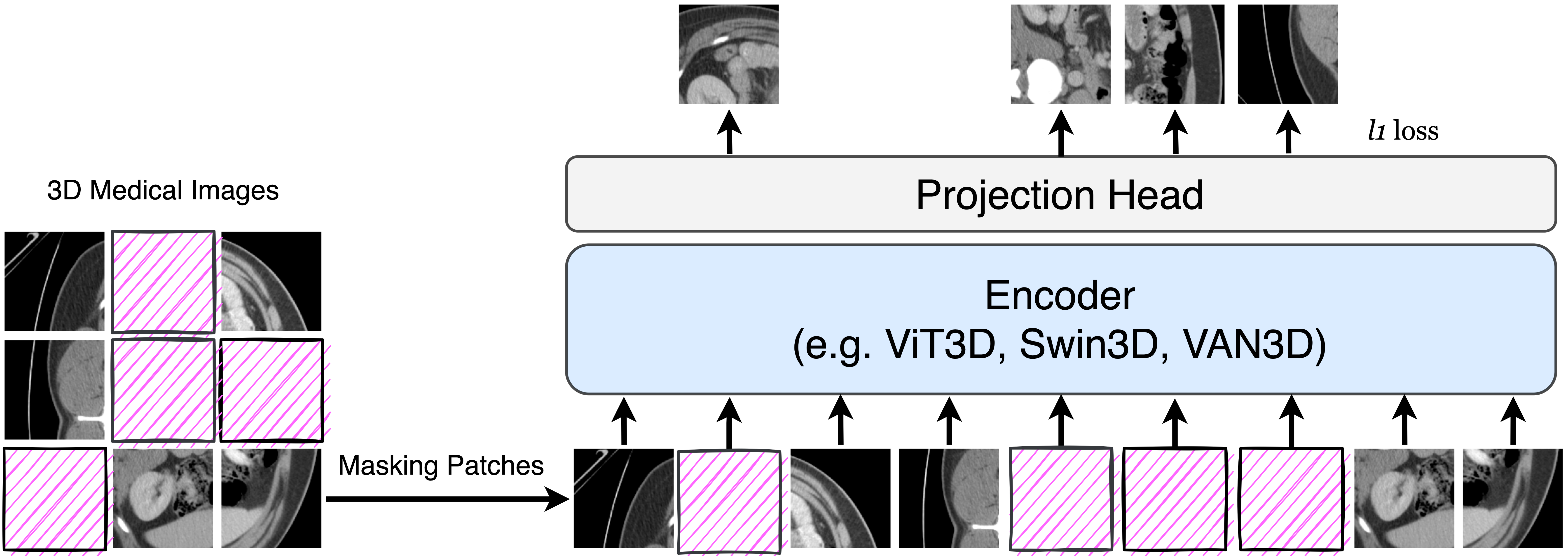}
    \caption{The SimMIM architecture.}
    \label{fig:simmim}
  \end{subfigure}
  
  \caption{Illustration of different self-supervised learning methods for modeling 3D medical images.}
  \label{fig:main}
\end{figure*}
% }

As a viable alternative, self-supervised learning~\cite{Jaiswal2020ASO} obtains supervisory signals from the data itself, and has recently been shown to successfully address the appetite for data and to be capable of learning generalizable dense representations of the input. Among contemporary approaches, \textit{masked signal modeling} is one such learning task: masking a subset of input signals and attempting to forecast the masked signals. This paradigm has been extremely successful in NLP, since self-supervised learning algorithms based on the masked language modeling task have largely revolutionized the discipline~\cite{Devlin2019BERTPO,Radford2018GPT1,Radford2019GPT2,Tom2020GPT3}, demonstrating that giant models such as BERT~\cite{Devlin2019BERTPO} and GPT~\cite{Radford2018GPT1,Radford2019GPT2,Tom2020GPT3} can be learned on unlabeled text data and adapted to a wide variety of applications. More importantly, with the introduction of Vision Transformers (ViT)~\cite{Vaswani2017AttentionIA,Dosovitskiy2021AnII}, the architecture gap, where it was not intuitive to apply mask tokens\cite{Vaswani2017AttentionIA,Devlin2019BERTPO} using covolutions~\cite{LeCun1989BackpropagationAT}, is no longer an obstacle. Following this philosophy, latest approaches based on \textit{masked image modeling} (MIM) have demonstrated their efficacy in the development of scalable vision models~\cite{He2021MaskedAA,Xie2021SimMIMAS,Baevski2022data2vecAG}. Despite these accomplishments, masked image modeling based algorithms have received little attention in medical imaging modeling, and their applicability has not been thoroughly investigated. Naturally, we wonder \textit{whether masked image modeling will advance 3D medical imaging analysis as well}. In this work, we aim to address this question from the following attempts:
\begin{itemize}
    \item Contrastive learning~\cite{Becker1992SelforganizingNN,Hadsell2006DimensionalityRB,Chen2020ASF} has been proven in a few studies to be capable of learning generic representations of medical images that leverage the downstream tasks such as 3D segmentation and classification~\cite{Taleb20203DSM,Azizi2021BigSM,Tang2021SelfSupervisedPO}. It is worthwhile to compare masked image modeling to contrastive learning approaches (see \cref{fig:simclr} for illustration) on medical images.
    \item Natural images are raw, low-level signals with a significant degree of spatial redundancy; restoring some missing patches can be accomplished by directly copying surrounding patches with little high-level understanding of the objects and sceneries~\cite{He2021MaskedAA}. Particularly for certain CT/MRI scans with solid tumors, the majority of background tissues are comparable, making it even more difficult for the model to learn useful features about the lesion regions. As a result, we assess several masking strategies (masked patch size and masking ratio) in order to determine the most efficient way that promotes holistic comprehension beyond low-level data while avoiding excessive attention to features such as texture and materials.
    \item In practice, medical image analysis is utilized in a variety of contexts with varying amounts of annotated data, accessible unlabeled data, and even image resolutions. As a result, it is also vital for us to extensively analyze how these elements affect the pertaining as well as the performance on downstream tasks.
\end{itemize}
This paper investigates how masked image modeling based self-supervised learning can be utilized to improve 3D medical image analysis. It does so by conducting extensive experiments on two real-world benchmark datasets: multi-organ segmentation\footnote{https://www.synapse.org/\#!Synapse:syn3193805/wiki/89480} and brain tumor segmentation~\cite{Simpson2019ALA}. Our experimental results demonstrate that masked image modeling is advantageous for modeling 3D medical images by significantly speeding up training convergence (\eg~at most 1.4$\times$ training cost saving to reach the same dice score) and ultimately improved downstream performance (\eg~over 5$\%$ improvements on both segmentation without any hyperparameter tuning).

% \comment{
\begin{figure*}[htb]
  \centering
  \includegraphics{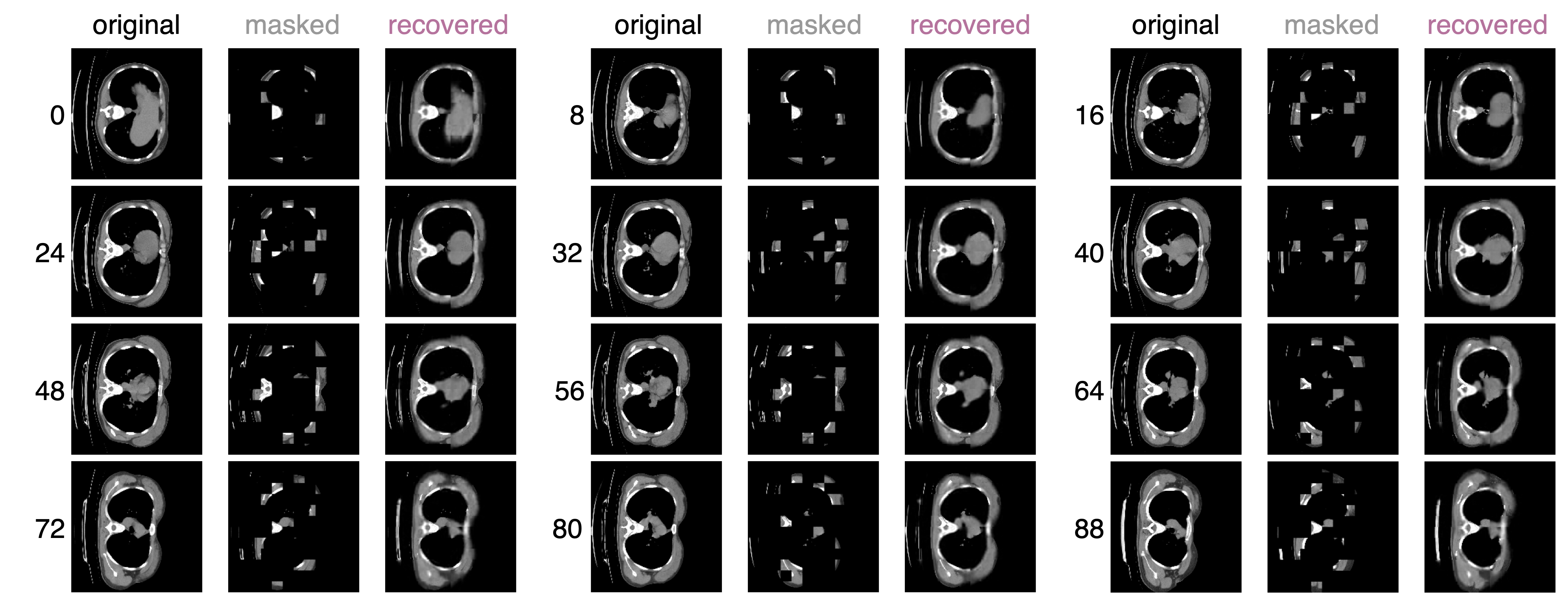}
  \caption{Example results of one CT scan from TCIA-COVID19~\cite{Harmon2020ArtificialIF} \textit{validation} set. As the original images are all 3D volumes, we show the reconstructed images in the form of slices, where the indexing number represents the depth. For each triplet, we show the ground truth (left), the masked image (middle), and the SimMIM~\cite{Xie2021SimMIMAS} reconstruction (right). In this case, a ViT-Base backbone is applied for the encoder, the masked patch size is 16 (for all dimensions), and the masking ratio is 75$\%$ following~\cite{Xie2021SimMIMAS}. }
  \label{fig:lung_recon_simmim}
\end{figure*}

\begin{figure*}[htb]
  \centering
  \includegraphics{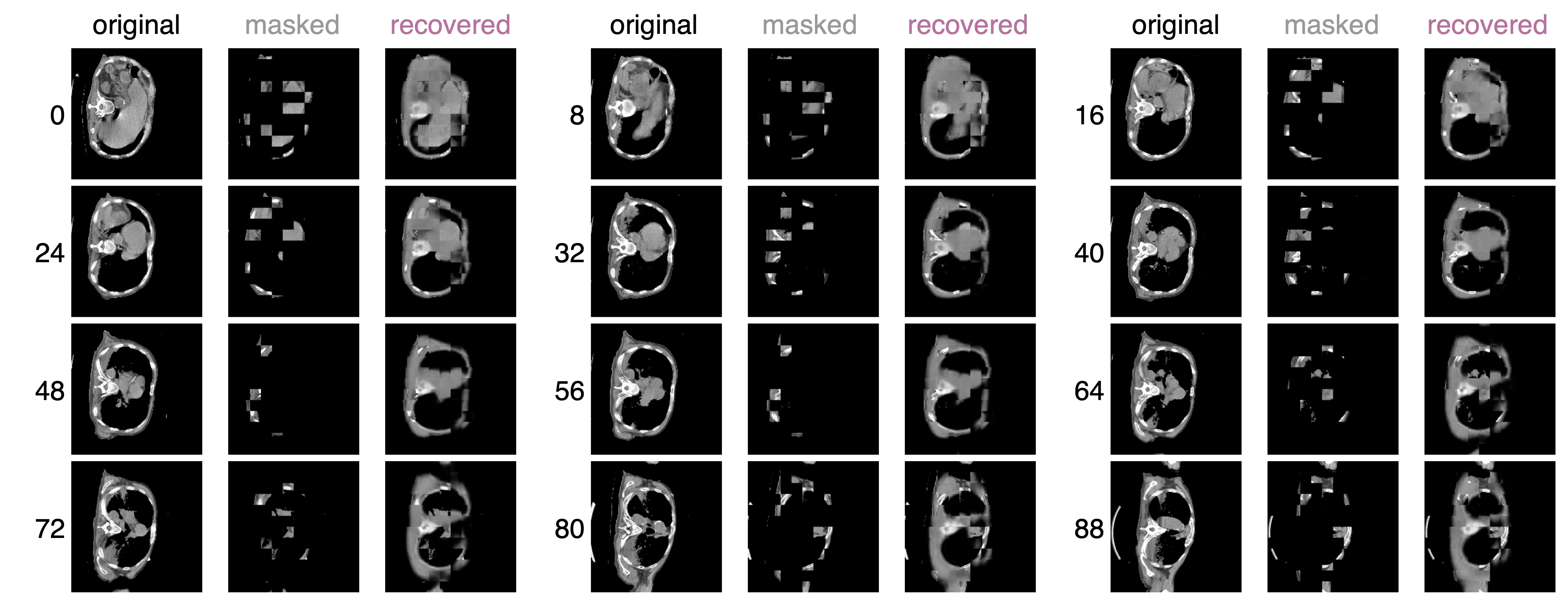}
  \caption{Example results of another CT scan from TCIA-COVID19 \textit{validation} set. Same as \cref{fig:lung_recon_simmim}, we show the reconstructed images slice by slice, where the indexing number represents the depth. For each triplet, we show the ground truth (left), the masked image (middle), and the MAE~\cite{He2021MaskedAA} reconstruction (right). In this case, a ViT-Large is applied as the encoder backbone, the masked patch size is 16 (for all dimensions), and the masking ratio is 75$\%$ following~\cite{He2021MaskedAA}.}
  \label{fig:lung_recon_finetune}
\end{figure*}
% }

%-------------------------------------------------------------------------
\section{Related Work}
\label{sec:relatedwork}

\paragraph{Transfer Learning in Medical Image Analysis.} Transfer learning from natural images is extensively utilized in medical image analysis~\cite{Liu2020ADL,McKinney2020InternationalEO}, regardless disparities in image statistics, scale, and task-relevant characteristics. Raghu \etal~\cite{Raghu2019TransfusionUT} and \cite{Azizi2021BigSM} showed that transfer learning from ImageNet can accelerate convergence on medical images, which is especially useful when the medical image training data is limited. Transfer learning using domain-specific data can also assist in resolving the domain disparity issue. For instance, \cite{Chen2019Med3DTL,Liang2020ATL} indicate improved performance following pretraining on labeled data from the same domain. However, this strategy is frequently impractical for a variety of medical scenarios requiring labeled data that is costly and time-consuming to gather. Recent improvements in self-supervised learning offer a viable alternative, allowing for the utilization of unlabeled medical data, which is massive and commonly more accessible.

\paragraph{Masked Image Modeling.} Masked image modeling is a self-supervised learning method that learns representations via masking-corrupted images. It evolved in line with the MLM task in NLP but remained out of the mainstream for a long period. DAE~\cite{Vincent2010StackedDA} is a pioneering work in this domain, presenting masking as a noise type. The context encoder~\cite{Pathak2016ContextEF} predicts the missing pixels by inpainting a large rectangular area of the source images. Recent techniques~\cite{Chen2020GenerativePF,Dosovitskiy2021AnII,Bao2021BEiTBP} based on Transformers~\cite{Vaswani2017AttentionIA} are motivated by the success of NLP. iGPT~\cite{Chen2020GenerativePF} groups pixel values into different clusters and classify unknown pixels. The ViT study~\cite{Dosovitskiy2021AnII} investigates masked patch prediction for self-supervised learning by predicting the mean color of images. BEiT~\cite{Bao2021BEiTBP} recently used a dVAE network to tokenize and forecast pixel values into discrete numbers~\cite{Oord2017NeuralDR,Ramesh2021ZeroShotTG}. More importantly, MAE~\cite{He2021MaskedAA} adheres the spirit of raw pixel restoration, demonstrating for the first time that masking a high proportion of the input images can yield a non-trivial and meaningful self-supervisory task. It adopts a design of autoencoder with a lightweight decoder, which reduces training costs even more. SimMIM~\cite{Xie2021SimMIMAS} takes it a step further and substitutes the entire decoder with a single linear projection layer, resulting in comparable results. The very recent approaches such as data2vec~\cite{Baevski2022data2vecAG} and CAE~\cite{Chen2022ContextAF} make predictions in the latent representation space from the visible patches to the masked patches, attempting to make MIM a more universal framework for self-supervised learning. Nonetheless, the techniques described above have only been shown to be useful for natural images modeling. In this work, we aim to investigate whether MIM approaches can also advance 3D medical image analysis.

\paragraph{Self-Supervised Learning.} Early work in self-supervised learning focuses on learning representations from unlabeled data so that a low-capacity classifier can achieve high accuracy using these embeddings~\cite{Doersch2015UnsupervisedVR,Wang2015UnsupervisedLO,Noroozi2016UnsupervisedLO,Zhang2016ColorfulIC,Pathak2017LearningFB,Gidaris2018UnsupervisedRL}. For years, contrastive learning~\cite{Becker1992SelforganizingNN,Hadsell2006DimensionalityRB,Wu2018UnsupervisedFL,Oord2018RepresentationLW,He2020MomentumCF,Chen2020ASF} has received much interest as one of the most popular and widespread self-supervised learning strategies. It models image similarity and dissimilarity (or solely similarity~\cite{Grill2020BootstrapYO,Chen2021ExploringSS}) between two or more views, with data augmentation being crucial for contrastive and related approaches. Self-supervised learning has also been used in the medical field, according to several previous literature. Domain-specific pretext tasks~\cite{Spitzer2018ImprovingCS,Bai2019SelfSupervisedLF,Zhuang2019SelfsupervisedFL,Zhu2020RubiksCA}, for example, have been studied, while other work~\cite{Liu2019AlignAA,He2020SampleEfficientDL,Zhou2020ComparingTL,Li2021ImbalanceAwareSL} focuses on tailoring contrastive learning to medical data. Taleb \etal~\cite{Taleb20203DSM}, in particular, examine a range of self-supervised learning strategies for 3D medical imaging in depth. MICLe~\cite{Azizi2021BigSM} demonstrates that a model pretrained on ImageNet can also advance Dermatology image classification. Tang \etal~\cite{Tang2021SelfSupervisedPO} further combines inpainting~\cite{Pathak2016ContextEF} with contrastive learning for medical segmentation. Despite the fact that all of these methods have showed promise in medical imaging, masked image modeling-based methods have yet to be substantially investigated in this discipline.

% Update the cvpr.cls to do the following automatically.
% For this citation style, keep multiple citations in numerical (not
% chronological) order, so prefer \cite{Alpher03,Alpher02,Authors14} to
% \cite{Alpher02,Alpher03,Authors14}.

%------------------------------------------------------------------------
\section{Approach}
\label{sec:approach}

Masked image modeling approaches, in general, mask out a portion of input images or encoded image tokens and encourage the model to recreate the masked area. Many extant MIM models employ an encoder-decoder design followed by a projection head, such as BEiT~\cite{Bao2021BEiTBP} and MAE~\cite{He2021MaskedAA}. The encoder aids in the modeling of latent feature representations, while the decoder aids in the resampling of latent vectors to original images. The encoded or decoded embeddings will subsequently be aligned with the original signals at the masked area by a projection head. Notably, the decoder component has been suggested to be designed in a lightweight manner in order to minimize training time. A lightweight decoder, in our experience, not only reduces computing complexity but also increases the encoder's ability to learn more generalizable representations that the decoder can easily grasp, translate and convey. As a result, while the encoder is more important (only encoder would be inherited for finetuning), methods like SimMIM~\cite{Xie2021SimMIMAS} simplifies the architecture even more by obviating the entire decoder with a single projection layer. In this work, we thoroughly investigate the effectiveness of different MIM models on 3D medical imaging data. The following components provide more details:

\subsection{Masking Strategies} 
Following ViT\cite{Dosovitskiy2021AnII}, an image is divided into regular non-overlapping patches (\eg~a 96$\times$ 96$\times$ 96 3D volume will be divided into 216 patches of 16$\times$ 16$\times$ 16 smaller volumes), which are often considered as the basic processing units of vision Transformers. Multiple random masking methods have been proposed in the previous literature: 1) InPainting~\cite{Pathak2016ContextEF} introduced a central region masking strategy; 2) BEiT~\cite{Bao2021BEiTBP} proposed a complex block-wise masking strategy; 3) most recent approaches such as MAE~\cite{He2021MaskedAA} and SimMIM~\cite{Xie2021SimMIMAS} followed a more straightforward uniformly random masking method at patch-level while investigating different masked patch sizes and masking ratios (see \cref{fig:mae} and \cref{fig:simmim}, respectively). Many random masking schemes are also patch-based since it is more convenient to operate masking on a patch-by-patch basis, where a patch is either fully visible or masked. As demonstrated by these works, the \textit{uniformly} random sampling with a \textit{high} masking ratio effectively eliminates redundancy, resulting in a self-supervisory task that cannot be easily solved by extrapolation from visible neighboring patches. Meanwhile, a potential center bias (\ie~more masked patches near the image center) is avoided by the uniform distribution. Finally, the sparse input allows for the development of an efficient encoder, which will be discussed next. In this work, we also use the random patch masking approach for simplicity and efficacy.

\subsection{Encoders} 
Encoders are responsible for modeling latent feature representations of the masked patches, which are then utilized to forecast the original signals in the masked area. The learned encoder should be capable of adapting to a wide range of vision tasks. We consider a variety of architectures in this paper, including two fundamental vision Transformer architectures: vanilla ViT~\cite{Vaswani2017AttentionIA,Dosovitskiy2021AnII} and SwinTransformer~\cite{Liu2021SwinTH}, as well as one attentional visual network VAN~\cite{Guo2022VisualAN}, which inherits the attention mechanism to derive hierarchical representations similar to SwinTransformer but using pure convolutions. All models are reimplemented to 3D versions in order to accommodate the 3D volume data. We simply refer to these models as ViT3D, SwinTransformer3D and VAN3D.

\subsection{Decoders}
For methods that follow an auto-encoder design to reconstruct the image, the decoder takes the entire collection of encoded tokens, including 1) encoded visible patches and 2) mask tokens. Each randomly initialized mask token is a learnable vector that is jointly optimized to reveal the masked patches. The absolute positional embeddings~\cite{Vaswani2017AttentionIA} or relative positional embeddings~\cite{Liu2021SwinTH} are also applied to these mask tokens corresponding to the backbone architecture. Additionally, all the masked patches are invisible to the encoder and only the decoder can see all tokens. As proved in~\cite{He2021MaskedAA}, this can save more computation and memory while not interfering with training. Meanwhile, the decoder backbones are independent from the encoder backbones, which are likewise optional (see \cref{fig:mae}). By default, we follow~\cite{He2021MaskedAA} and use another series of Transformer blocks for decoding. 

\subsection{Reconstruction Target}
\paragraph{Raw voxel value prediction.} For 3D medical image, reconstructing the inputs by estimating the raw voxel values for each mask token is simple and intuitive. The distance between recovered and original images in voxel space can be computed using a loss function of either $l_1$ loss or $l_2$ loss. Furthermore, the loss is only computed on masked patches, preventing the model from engaging in self-reconstruction, which might potentially dominate the learning process and ultimately impede knowledge learning. Notably, most vision Transformer topologies will downsample the original image resolution. For 3D medical images, a 96$\times$ volume resolution will be downsampled to 9$\times$ (\ie~1*9*9*9$\approx$768 using ViT-Base) and 3$\times$ using SwinTransformer or VAN. Therefore, for vanilla ViT, we apply a single linear projection layer to transform the latent embeddings to the original voxel space; for SwinTransformer and VAN, we apply two-layers convolutional transpose to upsample the compressed embeddings to the original resolution. See \cref{fig:lung_recon_simmim} and \cref{fig:lung_recon_finetune} for the reconstruction of 3D lung CT scans from TCIA-COVID19 using SimMIM~\cite{Xie2021SimMIMAS} and MAE~\cite{He2021MaskedAA}, respectively.
\paragraph{Other predictions.} Many earlier studies transform masked signals to clusters or classes rather than raw pixel values. For example, iGPT~\cite{Chen2020GenerativePF} uses \textit{k}-means to divide the RGB values into 512 clusters and encourage the model to predict which cluster each pixel belongs to. BEiT~\cite{Bao2021BEiTBP} employs a discrete VAE (dVAE) to convert image patches to discrete tokens. The prediction objective is then based on the token identity. Medical images, on the other hand, are often sparse, and voxel values are not scale intensive. The fine-grained texture or materials information may be lost by replacing the original signals with a discrete class target. As a result, we concentrate on predicting raw voxel values in this work for the sake of simplicity and robustness.

%-------------------------------------------------------------------------
\begin{table*}[!htb]
\resizebox{\linewidth}{!}{%
\begin{tabular}{@{}llcccccccccccccc@{}}
\toprule
\multirow{2}{*}{Methods} & \multirow{2}{*}{Backbones} & \multicolumn{13}{c}{Multi-Organ Sementation} & \multirow{2}{*}{Avg. $\uparrow$} \\ \cmidrule(l){3-15}
 &  & Spleen & RKid & LKid & Gall & Eso & Liv & Sto & Aor & IVC & Veins & Pan & RAG & LAG &  \\ \midrule
\multirow{2}{*}{\begin{tabular}[c]{@{}l@{}}Sup. baseline~\cite{Hatamizadeh2022UNETRTF}\\ our impl.\end{tabular}} & ViT3D-B~\cite{Dosovitskiy2021AnII} & 0.8902 & 0.8926 & 0.8769 & 0.4763 & 0.4891 & 0.9447 & 0.7475 & 0.8207 & 0.773 & 0.6175 & 0.6442 & 0.5663 & 0.4699 & 0.7084 \\
& ViT3D-L & 0.8993 & 0.9018 & 0.8859 & 0.4813 & 0.4942 & 0.9543 & 0.7553 & 0.8292 & 0.7810 & 0.6239 & 0.6508 & 0.5721 & 0.4749 & 0.7157 \\
 & Swin3D-T~\cite{Liu2021SwinTH} & 0.8638 & 0.8661 & 0.8508 & 0.4622 & 0.4746 & 0.9166 & 0.7255 & 0.7963 & 0.7500 & 0.5992 & 0.6252 & 0.5496 & 0.4560 & 0.6874 \\
 & Swin3D-S & 0.8792 & 0.8815 & 0.8661 & 0.4704 & 0.4831 & 0.9331 & 0.7383 & 0.8107 & 0.7635 & 0.6099 & 0.6363 & 0.5594 & 0.4641 & 0.6997 \\
 & Swin3D-B & 0.8852 & 0.8876 & 0.8721 & 0.4737 & 0.4864 & 0.9395 & 0.7434 & 0.8162 & 0.7687 & 0.6141 & 0.6407 & 0.5632 & 0.4673 & 0.7045 \\
 & VAN3D-S~\cite{Guo2022VisualAN} & 0.8572 & 0.8595 & 0.8444 & 0.4587 & 0.4711 & 0.9098 & 0.7198 & 0.7904 & 0.7444 & 0.5946 & 0.6204 & 0.5454 & 0.4525 & 0.6822 \\
 & VAN3D-B & 0.8813 & 0.8837 & 0.8682 & 0.4716 & 0.4843 & 0.9354 & 0.7401 & 0.8126 & 0.7653 & 0.6114 & 0.6379 & 0.5607 & 0.4653 & 0.7014 \\ \midrule
\multirow{2}{*}{SimCLR~\cite{Chen2020ASF}} & ViT3D-B~\cite{Dosovitskiy2021AnII} & 0.9110 & 0.9135 & 0.8974 & 0.4875 & 0.5007 & 0.9669 & 0.7650 & 0.8399 & 0.7912 & 0.6320 & 0.6594 & 0.5795 & 0.4810 & 0.7249 \\
 & ViT3D-L & 0.9279 & 0.9304 & 0.9141 & 0.4965 & 0.5099 & 0.9849 & 0.7792 & 0.8556 & 0.8058 & 0.6437 & 0.6716 & 0.5904 & 0.4899 & 0.7385 \\ \midrule
\multirow{2}{*}{MAE~\cite{He2021MaskedAA}} & ViT3D-B~\cite{Dosovitskiy2021AnII} & 0.9488 & 0.9502 & 0.9341 & 0.5066 & 0.521 & 0.9863 & 0.7969 & 0.8742 & 0.8248 & 0.6589 & 0.6868 & 0.6052 & 0.5010 & \underline{0.7534} \\
 & ViT3D-L & 0.9541 & 0.9566 & 0.9399 & 0.5105 & 0.5243 & 0.9878 & 0.8012 & 0.8797 & 0.8285 & 0.6618 & 0.6905 & 0.6070 & 0.5037 & \underline{0.7574} \\ \midrule
\multirow{6}{*}{SimMIM~\cite{Xie2021SimMIMAS}} & ViT3D-B~\cite{Dosovitskiy2021AnII} & 0.9520 & 0.9545 & 0.9378 & 0.5194 & 0.5232 & 0.9875 & 0.7995 & 0.8776 & 0.8267 & 0.6605 & 0.6890 & 0.6076 & 0.5126 & \underline{0.7575} \\
 & ViT3D-L & 0.9556 & 0.9582 & 0.9414 & 0.5206 & 0.5352 & 0.9898 & 0.8025 & 0.8811 & 0.8298 & 0.6649 & 0.6916 & 0.6088 & 0.5045 & \textbf{0.7603} \\
 & Swin3D-T~\cite{Liu2021SwinTH} & 0.9157 & 0.9182 & 0.9021 & 0.4900 & 0.5032 & 0.9719 & 0.7690 & 0.8443 & 0.7952 & 0.6352 & 0.6628 & 0.5826 & 0.4834 & 0.7288 \\
 & Swin3D-S & 0.9319 & 0.9344 & 0.9181 & 0.4987 & 0.5121 & 0.9891 & 0.7826 & 0.8593 & 0.8093 & 0.6465 & 0.6745 & 0.5929 & 0.4920 & 0.7416 \\
 & Swin3D-B & 0.9387 & 0.9413 & 0.9248 & 0.5023 & 0.5159 & 0.9963 & 0.7883 & 0.8656 & 0.8152 & 0.6512 & 0.6794 & 0.5973 & 0.4956 & 0.7471 \\
 & VAN3D-S~\cite{Guo2022VisualAN} & 0.9090 & 0.9115 & 0.8955 & 0.4864 & 0.4995 & 0.9648 & 0.7634 & 0.8382 & 0.7894 & 0.6306 & 0.6579 & 0.5784 & 0.4799 & 0.7234 \\
 & VAN3D-B & 0.9350 & 0.9375 & 0.9211 & 0.5003 & 0.5138 & 0.9924 & 0.7852 & 0.8621 & 0.8119 & 0.6486 & 0.6767 & 0.5949 & 0.4936 & 0.7441 \\ \bottomrule
\end{tabular}%
}
\caption{Main results on multi-organ segmentation task. All models are pretrained on a combination of BTCV and TCIA-COVID19~\cite{Harmon2020ArtificialIF} datasets. The BTCV \textit{validation} set is utilized for validation consistently. }
\label{tab:btcv}
\end{table*}

%-------------------------------------------------------------------------
\begin{figure}
  \centering
  \includegraphics[width=\linewidth]{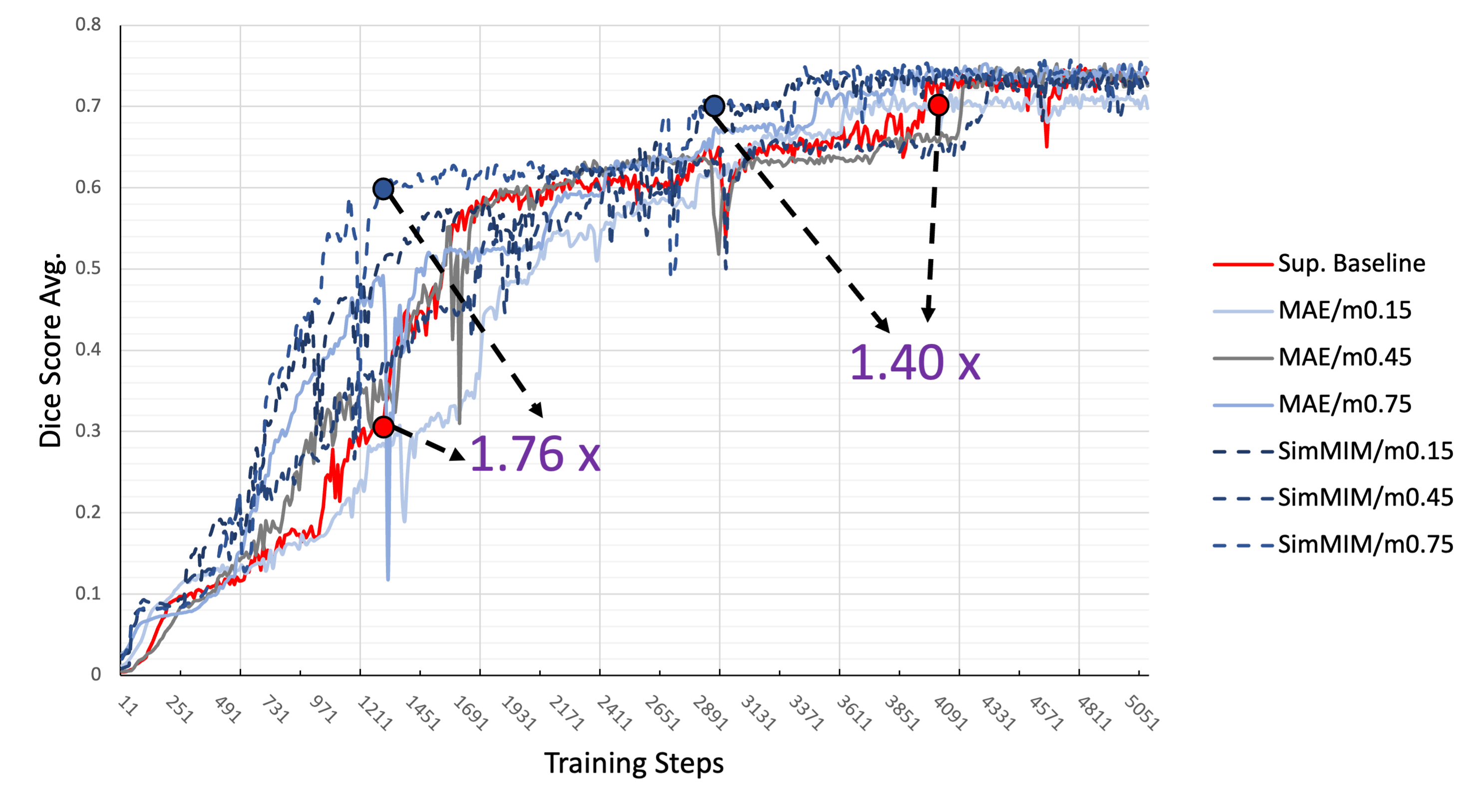}
  \caption{An illustration of how MIM pre-training advances the downstream supervised fine-tuning. We compare the average dice score on validation set between supervised baseline and different MIM methods using different masking ratios across training steps. Masked image modeling pre-training can significantly save training costs and generate better performance.}
  \label{fig:convergence}
\end{figure}

%------------------------------------------------------------------------
\section{Experiments on 3D Segmentation}
\label{sec:experiments}

We evaluate masked image modeling methods on two separate 3D segmentation tasks that involve both CT and MRI imaging modalities.
\paragraph{BTCV/CT.} The BTCV dataset\footnote{https://www.synapse.org/\#!Synapse:syn3193805/wiki/89480} comprises of 30 participants who had abdominal CT scans with 13 organs annotated by interpreters at Vanderbilt University Medical Center under the supervision of clinical radiologists. Each CT scan was performed in the portal venous phase with contrast enhancement and comprised of 80 to 225 slices with 512$\times$512 pixels and a slice thickness of 1 to 6 mm. Each volume was pre-processed separately, with intensities in the range of [-175, 200] HU being normalized to [0, 1]. During pre-processing, all images are resampled to 1.5/2.0 mm (different resolutions for ablation study) isotropic voxel spacing. The multi-organ segmentation problem is formulated as a 13 class segmentation task with 1-channel input. We use the first 24 volumes for training and report on 6 validation images.

\paragraph{BraTS/MRI.} Brain tumor segmentation is another important task. This BraTS dataset~\cite{Simpson2019ALA} contains a training set of 387 multi-modal multi-site MRI data (FLAIR, T1w, T1gd, T2w) with ground truth labels of gliomas segmentation necrotic/active tumor and edema is used for brain tumor segmentation. The voxel spacing of MRI images in this task is 1.0$\times$1.0$\times$1.0 mm$^3$. The voxel intensities are pre-processed with z-score normalization. The problem of brain tumor segmentation is formulated as a 3 class segmentation task with 4-channel input. We report on 97 validation images.

\paragraph{TCIA-COVID19/CT.} This is a public dataset~\cite{Harmon2020ArtificialIF} consisting of unenhanced chest CTs of patients with COVID19 infections. There are 771 volumes collected from 661 patients in total. All images are unannotated. We utilize this dataset as an extra unlabeled dataset for self-supervised learning in ablation study. All models in \cref{tab:btcv} are pretrained using a combination of this dataset and BTCV. In ablation study, we also compare the performance between pretraining with and without this dataset.

\paragraph{Supervised Baselines.} UNETR~\cite{Hatamizadeh2022UNETRTF} is a \textit{U-shaped} encoder-decoder architecture for medical segmentation that employs a ViT as the encoder backbone and a convolutional upsampling decoder following U-Net~\cite{Ronneberger2015UNetCN} design. It is one of the SOTA models in the domain of medical imaging segmentation that incorporates the vision Transformers as the backbone. UNETR-Base represents a ViT-Base~\cite{Dosovitskiy2021AnII} is applied as the encoder backbone. We adopt UNETR-B as the default supervised baseline in our ablation study. For other backbones (SwinTransformer and VAN) that produce hierarchical features, we adopt UPerNet~\cite{Xiao2018UnifiedPP} as the decode head by default for downstream segmentation. 

\paragraph{Evaluation Metric} We use the Dice score to evaluate the accuracy of segmentation in our experiments. For a given semantic class, let $G_i$ and $P_i$ denote the ground truth and prediction values for voxel $i$. The Dice score is defined as:
\begin{equation}
Dice(G, P) = \frac{2\sum_{i}^{I}G_{i}P_{i}}{\sum_{i=1}^{I}G_{i} + \sum_{i=1}^{I}P_{i}}
\end{equation}

\paragraph{Implementation Settings.} All of the models are implemented in PyTorch\footnote{https://pytorch.org}. We use MONAI\footnote{https://github.com/Project-MONAI} for data transformations and loading. In our ablation studies, we use ViT-Base~\cite{Dosovitskiy2021AnII} as the default encoder backbone. For supervised baseline of organ segmentation, we employ a batch size of 4, the AdamW~\cite{Loshchilov2019DecoupledWD} optimizer, and a learning rate of 0.0003 with a weight decay of 0.05 (because ViT based architectures are huge and easily overfit) based on a linear warmup up to 300 epochs and cosine annealing scheduler. Training is conducted on a single NVIDIA A10G instance for a total of 3000 epochs. For brain tumor segmentation, the batch size is set to 8 as the training is conducted on 4 NVIDIA A10G GPUs for a total number of 1000 epochs. We use a 100 epochs linear warmup and the optimizer settings are compatible with organ segmentation. Our Appendix provides additional information~\cref{sec:appendix}.

%-------------------------------------------------------------------------
\begin{table}[]
\center
\resizebox{\linewidth}{!}{%
\begin{tabular}{@{}llcccc@{}}
\toprule
\multirow{2}{*}{Methods} & \multirow{2}{*}{Backbones} & \multicolumn{3}{c}{Brain Tumor Sementation} &  \multirow{2}{*}{Avg. $\uparrow$} \\ \cmidrule(l){3-5}
 &  & TC & WT & ET &  \\ \midrule
\multirow{2}{*}{\begin{tabular}[c]{@{}l@{}}Sup. baseline~\cite{Hatamizadeh2022UNETRTF}\\ our impl.\end{tabular}} & ViT3D-B~\cite{Dosovitskiy2021AnII} & 0.8162 & 0.8781 & 0.5734 & 0.7559 \\
 & ViT3D-L & 0.8178 & 0.8798 & 0.5745 & 0.7574 \\ \midrule
\multirow{2}{*}{SimCLR~\cite{Chen2020ASF}} & ViT3D-B~\cite{Dosovitskiy2021AnII} & 0.8360 & 0.8988 & 0.5869 & 0.7739 \\
 & ViT3D-L & 0.8313 & 0.8944 & 0.5842 & 0.7699 \\ \midrule
\multirow{2}{*}{MAE~\cite{He2021MaskedAA}} & ViT3D-B~\cite{Dosovitskiy2021AnII} & 0.8690 & 0.9340 & 0.6104 & \underline{0.8045} \\
 & ViT3D-L & 0.8723 & 0.9385 & 0.6130 & \underline{0.8079} \\ \midrule
\multirow{6}{*}{SimMIM~\cite{Xie2021SimMIMAS}} & ViT3D-B~\cite{Dosovitskiy2021AnII} & 0.8734 & 0.9394 & 0.6103 & \underline{0.8077} \\
 & ViT-L & 0.8738 & 0.9401 & 0.6141 & \textbf{0.8093} \\
 & Swin3D-S~\cite{Liu2021SwinTH} & 0.8428 & 0.9067 & 0.5922 & 0.7806 \\
 & Swin3D-B & 0.8556 & 0.9205 & 0.6013 & 0.7924 \\
 & VAN3D-B~\cite{Guo2022VisualAN} & 0.8406 & 0.9043 & 0.5907 & 0.7785 \\
 & VAN3D-L & 0.8522 & 0.9169 & 0.5989 & 0.7893 \\ \bottomrule
\end{tabular}%
}
\caption{Main results on brain tumor segmentation. All models are pretrained on BraTS~\cite{Simpson2019ALA} \textit{training} set without extra data source.}
\label{tab:brats}
\end{table}

%-------------------------------------------------------------------------
\subsection{Comparison among Different Approaches}
We begin by evaluating 1) how masked image modeling methods compare to contrastive learning approaches and 2) how different masked image modeling approaches perform in comparison to one another using MAE~\cite{He2021MaskedAA} and SimMIM~\cite{Xie2021SimMIMAS} and a conventional contrastive learning methodology SimCLR~\cite{Chen2020ASF}. We evaluate a range of encoder backbones with varying network sizes, including pure vision Transformer~\cite{Dosovitskiy2021AnII}, SwinTransformer~\cite{Liu2021SwinTH}, and visual attentional network (VAN)~\cite{Guo2022VisualAN}. For MAE, we use an 8-layer Transformer block with 512-d as the decoder; for SimMIM, we use a single linear layer as the projection head. We use a two-layer convolutional transpose as the projection head for pretraining and the UPerNet~\cite{Xiao2018UnifiedPP} for segmentation in both Swin3D and VAN3D. All other hyper-parameters were set identically in this investigation. Additionally, because the full 3D image volume is typically difficult to load directly into the GPU (memory explosion), we employ a sliding window training strategy~\cite{Peiris2021AVT,Hatamizadeh2022UNETRTF,Hatamizadeh2022SwinUS} in which the original image is divided into several (96$\times$96$\times$96) small 3D windows. For all ViTs, a patch size of 16 is utilized by default.

\cref{tab:btcv} demonstrates that masked image modeling approaches outperform contrastive learning methods in general, as both MAE~\cite{He2021MaskedAA} and SimMIM~\cite{Xie2021SimMIMAS} achieve an average dice score of around 0.752$\sim$0.758, while SimCLR achieves an average dice score of around 0.723, which is 4.5$\%$ lower than the best approach. The segmentation findings for BraTS in \cref{tab:brats} follow a similar pattern. The average dice score for masked image modeling approaches is somewhat greater than 0.80, however SimCLR~\cite{Chen2020ASF} obtains a dice value of 0.7739, which is 4.37$\%$ lower than the best approach comparable to \cref{tab:btcv}. Another note is that, despite the similarity of the two MIM techniques, SimMIM~\cite{Xie2021SimMIMAS} achieves slightly better performance than MAE~\cite{He2021MaskedAA}, as demonstrated by both \cref{tab:btcv} and \cref{tab:brats}. One explanation for this phenomena is because an efficient decoder (even a lightweight one) may be able to reconstruct the original image even if the encoder does not acquire generalizable representations, hence cyclically ease the motivation of encoder to learn more effective representations. Self-supervised learning's ultimate goal is always to learn effective and generalizable representations of the data rather than self-convergence only. In comparison, SimMIM~\cite{Xie2021SimMIMAS} employs an even lighter design by omitting the decoder entirely, which pushes the encoder to perform more complex reconstruction and learning tasks.

%\cref{tab:btcv} shows that masked image modeling approaches generally outperform contrastive learning method as both MAE~\cite{} and SimMIM~\cite{} achieve the best average dice score around 0.752$\sim$0.758 while SimCLR only achieves an average dice score around 0.723, which is 4.5$\%$ lower than the best approach. The brats segmentation results in \cref{tab:brats} also demonstrate the same trend. The average dice score of masked image modeling approaches are all slightly over 0.80 while the SimCLR~\cite{} only achieves a dice score of 0.7739 which is 4.37$\%$ lower than the best one similar to \cref{tab:btcv}. Another observation here is that even though two MIM approaches receive similar results, however, SimMIM~\cite{} achieves a slightly better performance than MAE~\cite{} shown by \cref{tab:btcv} and \cref{tab:brats}. One hypothesis of this phenomenon is that an effective decoder (even lightweight) could potentially restore the original image even if the encoder does not learn generalizable representations which cyclically ease the motivation of encoder to learn more effective representations. The ultimate goal of self-supervised learning is always to learn effective and generalizable representations of the data itself rather than self-convergence. Contrastively, SimMIM~\cite{} utilizes an even lighter design by completely removing the decoder part which forces the encoder to take on heavier reconstruction and learning task.

Additionally, masked image modeling approaches dramatically increase the training speed and reduce the cost, as seen by \cref{fig:convergence}. SimMIM based architectures can obtain a 1.76$\times$ better dice score at the 1.3k training step. Moreover, MIM based approaches can reach a dice score of 0.7 with 1.4$\times$ less training time than the training time required for supervised baseline.

%-------------------------------------------------------------------------
\begin{table}[]
\resizebox{\linewidth}{!}{%
\begin{tabular}{@{}l|c|c|c@{}}
\toprule
Methods & Masked patch size & Masking ratio & Dice score Avg. $\uparrow$ \\ \midrule
\multirow{11}{*}{MAE~\cite{He2021MaskedAA}} & 16 & 0.15 & 0.7156 \\
 & 16 & 0.30 & 0.7114 \\
 & 16 & 0.45 & 0.6896 \\
 & 16 & 0.60 & 0.7153 \\
 & 16 & 0.75 & 0.7183 \\ \cmidrule(l){2-4}
 & 24 & 0.15 & 0.6471 \\
 & 24 & 0.45 & 0.7123 \\
 & 24 & 0.75 & \textbf{0.7244} \\ \cmidrule(l){2-4}
 & 32 & 0.15 & 0.7065 \\
 & 32 & 0.45 & 0.7184 \\
 & 32 & 0.75 & 0.7048 \\ \midrule
\multirow{11}{*}{SimMIM~\cite{Xie2021SimMIMAS}} & 16 & 0.15 & 0.7144 \\
 & 16 & 0.30 & 0.7248 \\
 & 16 & 0.45 & 0.7227 \\
 & 16 & 0.60 & 0.7208 \\
 & 16 & 0.75 & 0.7249 \\ \cmidrule(l){2-4}
 & 24 & 0.15 & 0.7292 \\
 & 24 & 0.45 & 0.7278 \\
 & 24 & 0.75 & 0.7156 \\ \cmidrule(l){2-4}
 & 32 & 0.15 & \textbf{0.7471} \\
 & 32 & 0.45 & 0.7264 \\
 & 32 & 0.75 & 0.7245 \\ \bottomrule
\end{tabular}%
}
\caption{Ablation study of different masked patch size and masking ratio on multi-organ segmentation. The default backbone of ViT-B is applied as the UNETR encoder. Notably, in this table, we compare models that have been pretrained on the BTCV training set alone; no other datasets are used. }
\label{tab:btcv-ablation}
\end{table}

\begin{table}[]
\resizebox{\linewidth}{!}{%
\begin{tabular}{@{}l|c|c|c@{}}
\toprule
Methods & Masked patch size & Masking ratio & Dice score Avg. $\uparrow$ \\ \midrule
\multirow{11}{*}{MAE~\cite{He2021MaskedAA}} & 16 & 0.15 & 0.7864 \\
 & 16 & 0.30 & 0.7854 \\
 & 16 & 0.45 & 0.7902 \\
 & 16 & 0.60 & 0.7965 \\
 & 16 & 0.75 & \textbf{0.8045} \\ \cmidrule(l){2-4}
 & 24 & 0.15 & 0.7412 \\
 & 24 & 0.45 & 0.7947 \\
 & 24 & 0.75 & 0.8041 \\ \cmidrule(l){2-4}
 & 32 & 0.15 & 0.7823 \\
 & 32 & 0.45 & 0.7819 \\
 & 32 & 0.75 & 0.8041 \\ \midrule
\multirow{11}{*}{SimMIM~\cite{Xie2021SimMIMAS}} & 16 & 0.15 & 0.7818 \\
 & 16 & 0.30 & 0.7923 \\
 & 16 & 0.45 & 0.7945 \\
 & 16 & 0.60 & 0.8058 \\
 & 16 & 0.75 & \textbf{0.8077} \\ \cmidrule(l){2-4}
 & 24 & 0.15 & 0.7852 \\
 & 24 & 0.45 & 0.7654 \\
 & 24 & 0.75 & 0.7982 \\ \cmidrule(l){2-4}
 & 32 & 0.15 & 0.7985 \\
 & 32 & 0.45 & 0.7958 \\
 & 32 & 0.75 & 0.7986 \\ \bottomrule
\end{tabular}%
}
\caption{Ablation study on different masked patch sizes and masking ratios on brain tumor segmentation. Likewise, the pretraining data consists entirely of the BraTS dataset itself and the ViT-B is applied as the encoder backbone in UNETR for segmentation finetuning.}
\label{tab:brats-ablation}
\end{table}

\begin{table}[]
\resizebox{\linewidth}{!}{%
\begin{tabular}{@{}clcc@{}}
\toprule
\begin{tabular}[c]{@{}c@{}}Resolutions\\ (downsampled ratio) \end{tabular} & Pretrain data & Labeled ratio & Dice avg. \\ \midrule
(2.0x, 2.0x, 2.0x) & COVID19 + BTCV & $50\%$ & 0.6919 \\
(2.0x, 2.0x, 2.0x) & COVID19 + BTCV & $100\%$ & 0.7338 \\
(1.5x, 1.5x, 2.0x) & COVID19 + BTCV & $50\%$ & 0.7024 \\
(1.5x, 1.5x, 2.0x) & COVID19 + BTCV & $100\%$ & \textbf{0.7534} \\
(2.0x, 2.0x, 2.0x) & BTCV & $50\%$ & 0.6552 \\
(2.0x, 2.0x, 2.0x) & BTCV & $100\%$ & 0.7018 \\
(1.5x, 1.5x, 2.0x) & BTCV & $50\%$ & 0.6814 \\
(1.5x, 1.5x, 2.0x) & BTCV & $100\%$ & 0.7183 \\ \bottomrule
\end{tabular}%
}
\caption{We use MAE~\cite{He2021MaskedAA} (p/16 and m/75$\%$) as the backbone for this ablation study. Models are pretrained on a variety of different data sources with varying degrees of downsampling. Then the pretrained models are finetuned on multi-organ segmentation dataset with varying labeled data ratios. Each model is validated using the same BTCV \textit{validation} set.}
\label{tab:resolutions}
\end{table}

%-------------------------------------------------------------------------
\subsection{Masking Strategy}
Additionally, we investigate the effectiveness of different masked patch sizes and masking ratios on self-supervised learning performance. The performance of several MIM techniques at finetuning segmentation is summarized in \cref{tab:btcv-ablation} and \cref{tab:brats-ablation}. The following observations are made: i) Consistent with the original MAE literature~\cite{He2021MaskedAA}, we conclude that a \textit{higher} masking ratio is a non-trivial self-supervised learning job that would continually drive the model to build generalizable representations that can be transferred effectively to downstream tasks. For example, the best dice scores on multi-organ segmentation and brain tumor segmentation tasks are obtained when a masking ratio of 0.75 is used across multiple patch sizes (e.g., 0.7183 for patch size 16 in \cref{tab:btcv-ablation}, and 0.8041 for patch sizes 24 and 32 in \cref{tab:brats-ablation}). ii) A \textit{high} masking ratio combined with a \textit{small} patch size likewise results in a relatively good performance when used in conjunction with SimMIM~\cite{Xie2021SimMIMAS}, similar to MAE~\cite{He2021MaskedAA}. As demonstrated by \cref{tab:btcv-ablation} and \cref{tab:brats-ablation}, when the patch size is equal to 16, the models perform optimally with dice scores of 0.7249 and 0.8077, respectively. iii) However, as the patch size increases, the SimMIM~\cite{Xie2021SimMIMAS} method appears to be less sensitive to this masking ratio. For instance, when the patch size is 32, models can earn the highest dice score with a masking ratio of 0.15, the smallest possible masking ratio. One hypothesis is that medical images are typically raw, low-level signals with a large degree of spatial redundancy; recovering some missing patches can be performed by directly copying nearby patches with little comprehensive knowledge of the objects and surroundings. A single small masked patch is incapable of adequately masking complicated and intersecting structures or locations, but a high patch size may be able to hide more significant signals independently. As a result, a \textit{high} masking ratio for small patch sizes is more critical than a \textit{high} masking ratio for big patch sizes.

\subsection{Data \vs Resolutions \vs Labeled Ratio}
In this section, we analyze the results to address the following three questions: i) Does increasing the amount of pretraining data improve downstream performance? ii) How do different pretrained resolutions affect downstream knowledge transfer? And iii) how do masked image learning approaches improve performance when using varying amounts of labeled data? All pretraining in \cref{tab:resolutions} is based on the MAE~\cite{He2021MaskedAA} architecture, which utilizes a ViT-Base/16 as the backbone with a masking ratio of 75$\%$, as demonstrated in \cref{tab:btcv-ablation} and \cref{tab:brats-ablation}. Differently labeled ratios indicate that we employ a varying percentage of annotated BTCV CT scans (e.g., 50$\%$ = 12 images, 100$\%$ = 24 images) for downstream finetuning, whereas the \textit{validation} set of 6 images is consistent.

%In this section, we analyze the results to address three following questions: i) does more pretraining data give better downstream performance?; ii) how different pretrained resolutions affect the downstream knowledge transferring; and iii) how self-supervised learning approaches lift up the performance using different amounts of labeled data. All pretraining in \cref{tab:resolutions} are based on MAE~\cite{} architecture using a ViT-Base/16 as the backbone and a masking ratio of 75$\%$ as it's been proven reasonable in \cref{tab:btcv-ablation} and \cref{tab:brats-ablation}. Different labeled ratios represent that we use different percentage of annotated BTCV CT scans (\eg~50$\%$=12 images, $100\%$=24 images) for downstream finetuning while the \textit{validation} set that contains 6 images is consistent. 

In the majority of supervised learning cases, more training data results in improved performance. Given that the majority of medical images are similar from the bottom logic up, we ask if this holds true in the case of self-supervised learning, and in particular, how much benefits can be gained through \textit{size} of pretrain data when utilizing MIM for 3D medical analysis. We adopt multi-organ segmentation as the example downstream task and create two distinct training scenarios: one that uses both COVID19 and BTCV datasets and another that uses only BTCV. \cref{tab:resolutions} demonstrates the constant tendency that models trained on more plentiful pretrained data outperform models trained on less pretrained data (e.g., 0.7534$\rightarrow$0.7183: 4.9$\%$ improvements, 0.7338$\rightarrow$0.7018: 4.6$\%$ improvements). This advantage is even more pronounced at lower image resolutions, as 0.6919 is 5.6$\%$ more than 0.6552 when only half labeled data is used.

%Under most scenarios of supervised learning, more training data result in better performance. Considering most of the medical images are similar from the bottom logic, we also wonder does this still hold in the case of self-supervised learning, especially how much benefits can be brought by the \textit{size} of pretrain data using MIM for 3D medical analysis. We adopt multi-organ segmentation as the downstream task for examination and design two different training scenarios: one pretraining using both COVID19 and BTCV datasets while one pretraining uses only BTCV. \cref{tab:resolutions} demonstrate the consistent trends that more abundant pretrained data outperform the models pretrained on less data (\eg~0.7534$\rightarrow$0.7183: 4.9$\%$ improvements, 0.7338$\rightarrow$0.7018: 4.6$\%$ improvements). This improvement can be even more significant on lower image resolutions as 0.6919 is 5.6$\%$ higher than 0.6552 using only half labeled data. 

In \cref{tab:resolutions}, we also explore how different pretrained image resolutions affect the downstream task performance. Intuitively, a higher pretraining resolution should result in a better segmentation results~\cite{Azizi2021BigSM}, as the images contain more granular information. Here, we utilize different downsampled ratios to represent the degree to which the original signals are compressed in all dimensions for each volume. Specifically, a \textit{bilinear} interpolation function is used in conjunction with MONAI's\footnote{https://github.com/Project-MONAI} \textit{spacingd} transformations. As can be observed from \cref{tab:resolutions}, pretrained models with higher resolutions (1.5x, 1.5x, 2.0x) generally perform better than pretrained models with lower resolutions (2.0x, 2.0x, 2.0x). For instance, 0.7338 dice score is 2.7$\%$ lower than the one pretrained using the same data source and labeled ratio but using a greater resolution. 

In practical situation, the majority of the medical images, such as CT/MRI scans, are left unannotated due to the high cost of labeling. However, the public data is freely available and abundant, the aforementioned results illustrate once again that pre-training on large datasets followed by fine-tuning with small samples is feasible. It also demonstrates that masked image learning can significantly improve the downstream task performance in a variety of contexts.

%-------------------------------------------------------------------------
\section{Conclusion}
\label{sec:conclusion}
This paper demonstrates how masked image modeling approaches in self-supervised learning leverage 3D medical image modeling by conducting extensive experiments on two sample segmentation tasks. We show how masked image modeling outperforms traditional contrastive learning by speeding up convergence and greatly improving downstream task performance. We also show how masked image modeling approaches can be utilized to advance 3D medical image modeling in a variety of situations. However, the fact that almost all medical images are weakly labeled (\eg~as little as few lines of text for description) rather than entirely unannotated is an open question we would like to investigate further in the future. We are then interested on comparing self-supervised learning to supervised learning with limited supervisory signals. Finally, we remain curious to see how self-supervised learning is integrated in other more challenging downstream tasks.
% Most importantly, how to use these learned low-dimensional radiomic features as a clinical outcome endpoint and ultimately make a positive impact to patient healthcare is still a work in progress.

%%%%%%%%% REFERENCES
% \newpage
{\small
\bibliographystyle{ieee_fullname}
\bibliography{egbib}
}

%------------------------------------------------------------------------
% \newpage
\section{Appendix}
\label{sec:appendix}

\subsection{Experimental Settings}

\begin{table}[htb]
\resizebox{\linewidth}{!}{%
\begin{tabular}{l|l}
config & value \\ \toprule
optimizer & AdamW\cite{Loshchilov2019DecoupledWD} \\
base learning rate & 3e-4 \\
weight decay & 0.005 \\
optimizer momentum & beta1, beta2 = 0.9, 0.999 \\
batch size & 4 \\
learning rate schedule & linear warmup cosine annealing \\
warmup epochs & 300 \\
total epochs & 3000 \\
augmentation & RangeScaleIntensity
\end{tabular}%
}
\caption{Sup. baseline setting for BTCV.}
\label{tab:btcv-setting}
\end{table}

\begin{table}[htb]
\resizebox{\linewidth}{!}{%
\begin{tabular}{l|l}
config & value \\ \toprule
optimizer & AdamW\cite{Loshchilov2019DecoupledWD} \\
base learning rate & 3e-4 \\
weight decay & 0.005 \\
optimizer momentum & beta1, beta2 = 0.9, 0.999 \\
batch size & 2 \\
learning rate schedule & linear warmup cosine annealing \\
warmup epochs & 100 \\
total epochs & 1000 \\
augmentation & NormalizedIntensity
\end{tabular}%
}
\caption{Sup. baseline setting for BraTS.}
\label{tab:brats-setting}
\end{table}

\begin{table}[htb]
\resizebox{\linewidth}{!}{%
\begin{tabular}{l|l}
config & value \\ \toprule
optimizer & AdamW\cite{Loshchilov2019DecoupledWD} \\
base learning rate & 3e-4 \\
weight decay & 0.005 \\
optimizer momentum & beta1, beta2 = 0.9, 0.999 \\
batch size & 4 \\
learning rate schedule & linear warmup cosine annealing \\
warmup epochs & 300 \\
total epochs & 3000 \\
augmentation & RangeScaleIntensity
\end{tabular}%
}
\caption{Pretraining on CT 3D volumes.}
\label{tab:ct-setting}
\end{table}

\begin{table}[htb]
\resizebox{\linewidth}{!}{%
\begin{tabular}{l|l}
config & value \\ \toprule
optimizer & AdamW\cite{Loshchilov2019DecoupledWD} \\
base learning rate & 3e-4 \\
weight decay & 0.005 \\
optimizer momentum & beta1, beta2 = 0.9, 0.999 \\
batch size & 2 \\
learning rate schedule & linear warmup cosine annealing \\
warmup epochs & 100 \\
total epochs & 1000 \\
augmentation & RangeScaleIntensity
\end{tabular}%
}
\caption{Pretraining setting on MRI 3D volumes.}
\label{tab:mri-setting}
\end{table}

As can be seen from the reconstructed volumes, the large model has more restoration power than the tiny model, which supports the previous conclusion. The flattened dimensionality of 3D medical images is frequently very high, and a small model would unavoidably compress the original voxel space into a smaller voxel space, thereby losing a lot of information.

\begin{figure*}[]
  \centering
  \includegraphics{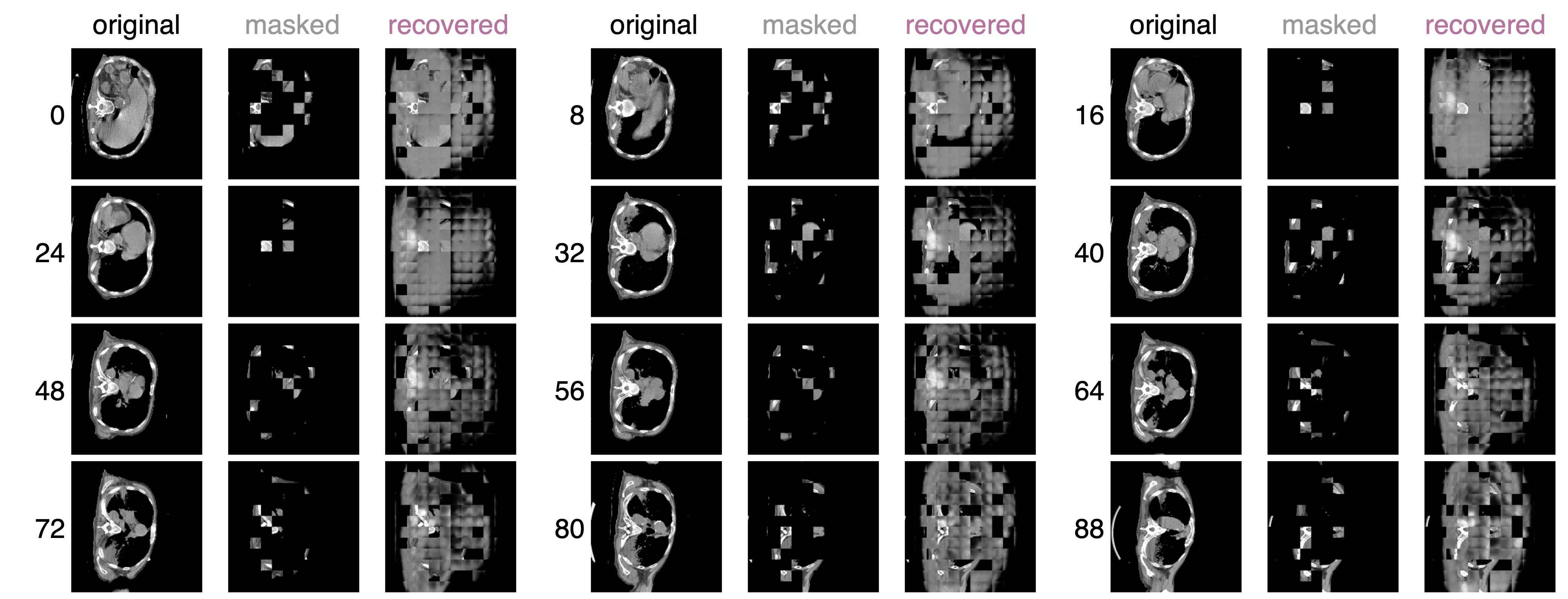}
  \caption{Example results of one CT scan from TCIA-COVID19. As the original images are all 3D volumes, we show the reconstructed images in a form of slices, where the indexing number represents the depth. For each triplet, we show the ground truth (left), the masked image (middle), and the SimMIM~\cite{Xie2021SimMIMAS} reconstruction (right). In this case, a ViT-Base backbone is applied for the encoder, the masked patch size is 16 (for all dimensions), and the masking ratio is 75$\%$ following~\cite{Xie2021SimMIMAS}. }
  \label{fig:lung_recon_base}
\end{figure*}

\begin{figure*}[]
  \centering
  \includegraphics{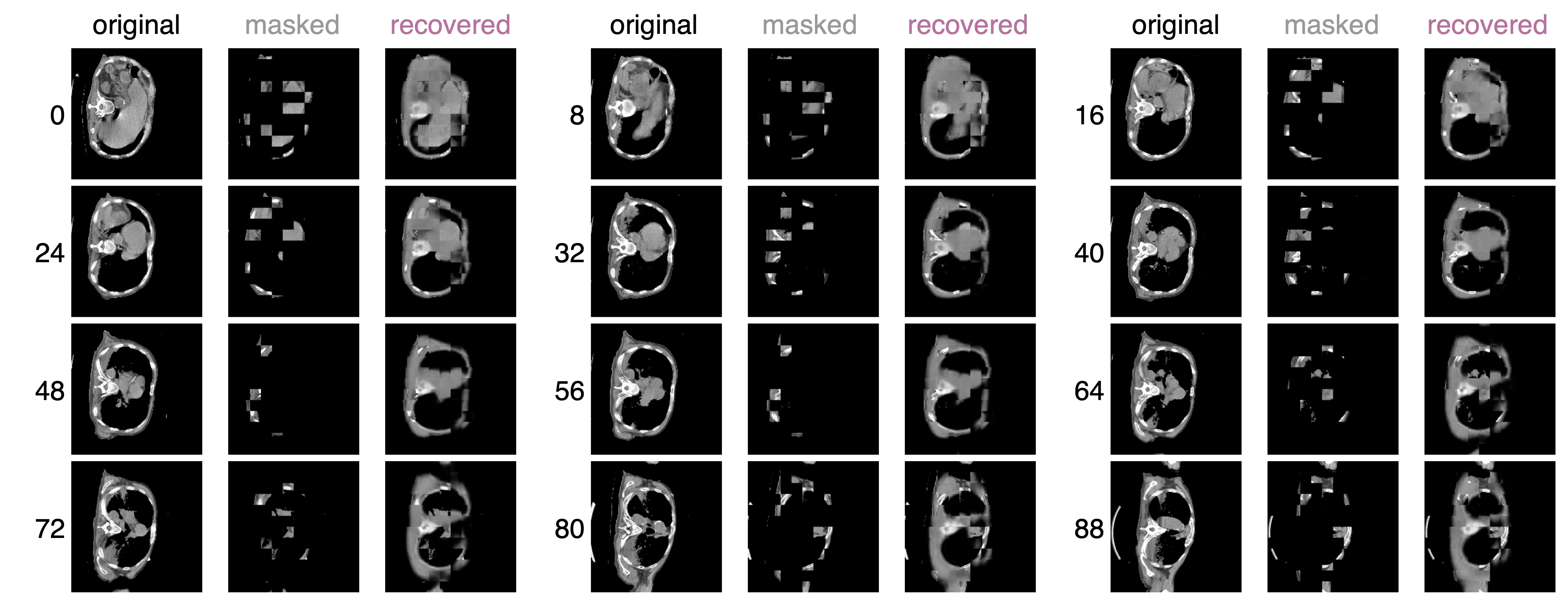}
  \caption{Example the same image as~\cref{fig:lung_recon_base}. As the original images are all 3D volumes, we show the reconstructed images in a form of slices, where the indexing number represents the depth. For each triplet, we show the ground truth (left), the masked image (middle), and the SimMIM~\cite{Xie2021SimMIMAS} reconstruction (right). In this case, a ViT-Large backbone is applied for the encoder. All rest settings are consistent with ~\cref{fig:lung_recon_base}. }
  \label{fig:lung_recon_large}
\end{figure*}

\end{document}